\documentclass{article}

\usepackage{arxiv}

\usepackage{url}            
\usepackage{booktabs}       
\usepackage{amsfonts}       
\usepackage{nicefrac}       
\usepackage{microtype}      

 \usepackage{hyperref}
 \hypersetup{colorlinks=true,linkcolor=citeblue,citecolor=citeblue,urlcolor=citeblue}

\usepackage{cite}
\usepackage{amsmath,amssymb,amsfonts}
\usepackage{cleveref}
\usepackage{subcaption}
\usepackage{makecell}

\usepackage[T1]{fontenc}
\usepackage{lmodern}

\usepackage{lipsum}
\usepackage{xcolor}
\usepackage{graphicx}
\usepackage{geometry}
\usepackage{array}
\usepackage{multirow}

\def\bng{\bngx}

\font\bngx=bang10

\def\*#1*#2{o\null{#2}{#1}}

\def\sh#1{\setbox0=\hbox{#1}%
     \kern-.02em\copy0\kern-\wd0
     \kern.04em\copy0\kern-\wd0
     \kern-.02em\raise.0433em\box0 }
\usepackage{xcolor}
\definecolor{citeblue}{RGB}{0,0,255}

\graphicspath{ {./images/} }

\title{Uddessho: An Extensive Benchmark Dataset for Multimodal Author Intent Classification in Low-Resource Bangla Language}

\author{
\\
\newline
\\
Fatema Tuj Johora Faria \textsuperscript{*},
Mukaffi Bin Moin,
Md. Mahfuzur Rahman,
Md Morshed Alam Shanto,
\\
Asif Iftekher Fahim, 
Md. Moinul Hoque
\\
\bigskip
\\
\\Ahsanullah University of Science and Technology, Dhaka, Bangladesh.
\\
\\
\\
\bigskip
*Corresponding author(s). E-mail(s): \texttt{\textcolor{blue}{fatema.faria142@gmail.com}}\\
Contributing authors: \texttt{\textcolor{blue}{mukaffi28@gmail.com}}; \texttt{\textcolor{blue}{mahim1066@gmail.com}}; \\\texttt{\textcolor{blue}{mdmorshedalamlipson@gmail.com}}; \texttt{\textcolor{blue}{fahimthescientist@gmail.com}}; \texttt{\textcolor{blue}{moinul.cse@aust.edu}}; 
\\ 
} 

\begin{document}

\maketitle
\abstract{
With the increasing popularity of daily information sharing and acquisition on the Internet, this paper introduces an innovative approach for intent classification in Bangla language, focusing on social media posts where individuals share their thoughts and opinions. The proposed method leverages multimodal data with particular emphasis on authorship identification, aiming to understand the underlying purpose behind textual content, especially in the context of varied user-generated posts on social media. Current methods often face challenges in low-resource languages like Bangla, particularly when author traits intricately link with intent, as observed in social media posts. To address this, we present the Multimodal-based Author Bangla Intent Classification (MABIC) framework, utilizing text and images to gain deeper insights into the conveyed intentions. We have created a dataset named ``Uddessho,'' comprising 3,048 instances sourced from social media. Our methodology comprises two approaches for classifying textual intent and multimodal author intent, incorporating early fusion and late fusion techniques. In our experiments, the unimodal approach achieved an accuracy of 64.53\% in interpreting Bangla textual intent. In contrast, our multimodal approach significantly outperformed traditional unimodal methods, achieving an accuracy of 76.19\%. This represents an improvement of 11.66\%. To our best knowledge, this is the first research work on multimodal-based author intent classification for low-resource Bangla language social media posts.}


\section{Introduction}
In the era of digital communication, understanding the underlying intentions behind textual content has become increasingly vital, especially in the context of social media where users express a wide range of thoughts and emotions. Multimodal social media platforms like Facebook, Instagram, and X (previously known as Twitter) \cite{intro_new} allow content creators to merge visual and textual components. The extensive use of text and images in social media posts underscores the necessity of accurately determining author intent within multimodal content, particularly in the context of the Bangla language. This poses a significant challenge for Natural Language Processing (NLP) in document understanding. Traditional approaches to intent classification \cite{intro_new1,intro_new2,intro_new3} primarily rely on text-based unimodal methods, which often struggle to fully capture the context and nuanced meanings inherent in multimodal posts. This limitation becomes particularly apparent when interpreting author intentions embedded within both visual and textual components.

This paper \cite{intro1} addressed Bangla intent classification in natural language understanding with BNIntent30, a Bengali dataset of 30 intent classes. The GAN-BnBERT approach, integrating Generative Adversarial BERT with contextual embeddings, outperformed existing models. However, the focus on unimodal text-based methods limited the model's ability to capture contextual information from visual cues in social media posts. In another study \cite{intro2}, a ensemble learning classifier was utilized for precise intent discernment in identifying hate speech on Tumblr. However, because it was limited to posts in English, it ignored subtleties in Bangla content, revealing a lack of linguistic inclusivity. Conversely, the paper \cite{intro3} on intent detection and slot filling in Sylheti and Bangla gave preference to unimodal text-based techniques and GPT-3.5 over JointBERT. However, the absence of investigation into multimodal techniques in this study limited the possibility of interpreting the author's aim in a more complex way.

Our research proposes a more comprehensive approach to author intent classification, combining textual and visual analysis to capture hidden meanings. We highlight the importance of MABIC over typical unimodal approaches for analyzing social media postings. By addressing the constraints of text-only techniques, we emphasize the value of multimodal methodologies in revealing the multifaceted characteristics of author intent. The contributions of this research paper can be summarized as follows:
\begin{itemize}
    \item[$\bullet$] We created the \textbf{``Uddessho''} or ``{\bng Ued/dshY}'' in Bangla, which means ``Intent'' in English, dataset for multimodal author intent classification, containing 3048 post instances across six categories: Informative, Advocative, Promotive, Exhibitionist, Expressive, and Controversial. The dataset is split into a training set with 2423 posts, a testing set with 313 posts, and a validation set with 312 posts, totaling 3048 posts.
    
    \item[$\bullet$] We investigated fusion techniques within the \textbf{MABIC framework}, such as Early and Late Fusion, which integrate features from multiple modalities. Our exploration showcased the superiority of multimodal approaches over text-based unimodal methods, underscoring the significance of incorporating diverse modalities for comprehensive author intent analysis.

        \item[$\bullet$] In our \textbf{qualitative error} investigation, We discovered that it was quite difficult to understand confusing textual and visual content. These challenges are especially noticeable when attempting to understand the author's hidden intentions, which are distinctive to the low resource Bangla language.
\end{itemize}

\section{Literature Review}
Julia et al. \cite{Rel1} examined text-image dynamics in social media, unveiling ``meaning multiplication.'' Their annotated dataset and fusion techniques highlighted the importance of understanding nuanced interactions between text and image. Additionally, their findings advocate for deeper exploration of divergent meanings in future investigations. Lu et al. \cite{Rel2}, on the other hand, proposed the MMIA approach, revolutionizing multimodal marketing intent analysis. Their method surpassed previous techniques in detecting intent within social news, outperforming SupDocNADE. Furthermore, MMIA excelled in topic identification, signifying a significant advancement in understanding and deciphering complex multimodal cues within marketing contexts. Shaozu et al. \cite{Rel3} introduced OCRBERT and VisualBERT for multimodal conversational intent classification in E-commerce. Both models, leveraging the MCIC dataset, outperformed baseline BERT in accuracy. Notably, VisualBERT showcased superior performance in capturing image regions, while OCRBERT emphasized the importance of OCR text integration for effective classification. Lastly, Mehedi et al. \cite{intro1} introduced BNIntent30, a Bengali intent classification dataset emphasizing linguistic diversity. They proposed GAN-BnBERT, integrating GAN with BERT to enhance classification. Experimental results showcased GAN-BnBERT's superiority. Although previous research has made considerable progress in unimodal and multimodal intent classification across multiple domains, there is still a significant gap in implementing these techniques to low-resource languages such as Bangla.

\section{Dataset Creation}\label{AA}
\subsection{\textbf{Data Collection}}
For our study, we manually collected a dataset of Bangla language social media posts from Facebook, X, and Instagram. We targeted popular accounts and groups with high engagement and diverse content to capture various author intents. Posts were selected based on the presence of both text and image, relevance to common Bangla social media topics, and diversity of expressed intents. Themes covered include personal updates, food reviews, sports, political news, entertainment, tourism, technology, pet, and promotional content.

\subsection{\textbf{Dataset Characteristics}}
The dataset comprises ``Image\_ID'' for image identification, ``Image\_Caption'' for associated text, and ``Intent\_Taxonomy'' for categorizing intentions. In our research, we defined six categories within the ``Uddessho'' (``{\bng Ued/dshY}'') dataset. Table \ref{tab:dataset_split} displays the statistics of the dataset. this research is publicly accessible through: \url{https://data.mendeley.com/datasets/mzxmt8tfjs/1}

\begin{enumerate}
\item[$\bullet$] \textbf{Advocative:} Posts advocating for a cause, figure, or idea, aiming to persuade others to support the advocated position.
\item[$\bullet$] \textbf{Promotive:}  Posts promoting events, products, or organizations to raise awareness and encourage participation.
\item[$\bullet$] \textbf{Exhibitionist:}  Posts showcasing personal identity, lifestyle, or belongings, such as selfies or pictures of pets.
\item[$\bullet$] \textbf{Expressive:} Posts expressing emotions, attachments, or admiration towards others.
\item[$\bullet$] \textbf{Informative:} Posts providing factual information about subjects or events, enriching understanding.
\item[$\bullet$] \textbf{Controversial:} Posts aiming to provoke debate or shock viewers by addressing taboo or contentious subjects.
\end{enumerate}

\begin{table*}
    \centering
    \caption{Distribution of Dataset Splits Across Different Intent Categories}
    \begin{tabular}{@{}l@{\hspace{5pt}}c@{\hspace{5pt}}c@{\hspace{5pt}}c@{}}
    \toprule
    \textbf{Intent\_Taxonomy} & \textbf{Train} & \textbf{Test} & \textbf{Validation} \\
    \midrule
    Informative & 514 & 67 & 67 \\
    Advocative & 386 & 49 & 49 \\
    Promotive & 315 & 43 & 42 \\
    Exhibitionist & 371 & 47 & 48 \\
    Expressive & 518 & 66 & 66 \\
    Controversial & 319 & 41 & 40 \\ 
    \bottomrule 
    \end{tabular}
    \label{tab:dataset_split}
\end{table*}

\subsection{\textbf{Annotation Guideline and Quality Maintenance}} 
 We recruited six annotators, aged (22-25) years, all undergraduates active on social media with a strong understanding of intention posts. They were compensated according to the Bangladesh wage payscale for their work. We provided annotators with clear guidelines for Multimodal Bangla Author Intent Classification. Aligning captions and images is essential for contextual comprehension, with any discrepancies documented for research. Our hired annotators classified postings based on language, visuals, and context, using a predefined taxonomy. They evaluated posts in their entirety, focusing on consistency while adhering to norms and conducting common validation tests. Each annotation includes transparent evidence of the rationale and decision-making process. Ongoing feedback sessions improve annotation skills and strengthen understanding of Bangla social media content intent. We achieved a Fleiss Kappa \cite{kappa} value of 0.84, suggesting strong agreement among annotators.

\section{Proposed Methodology}
\begin{figure}[htbp]
     \includegraphics[width=1\textwidth]{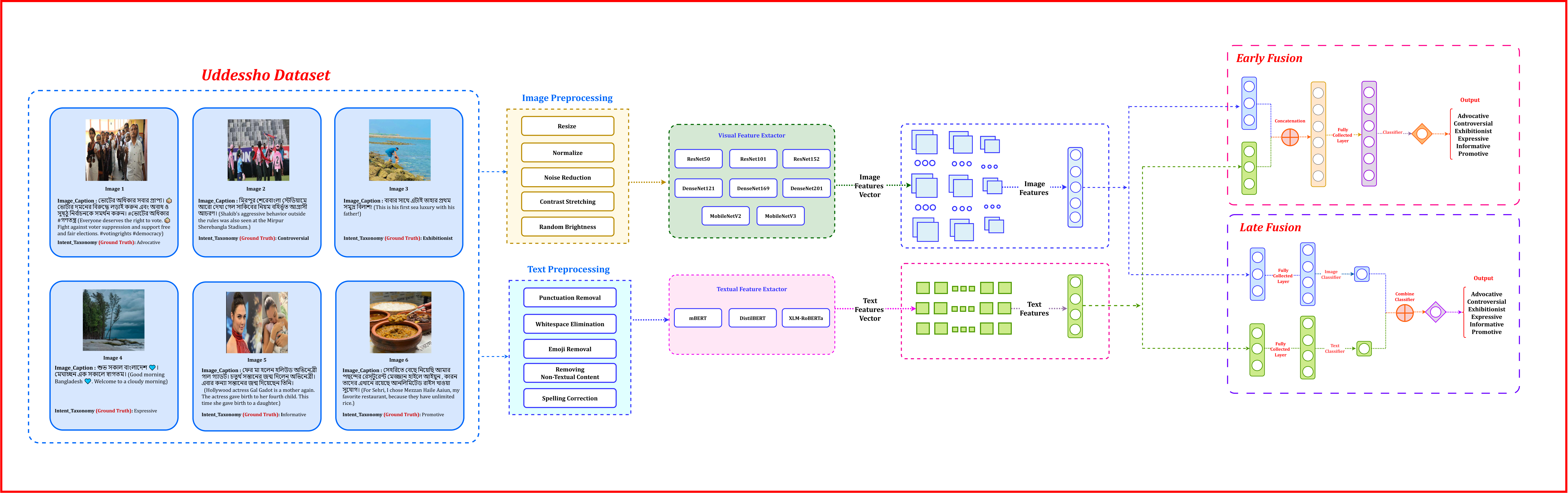}
    \caption{The diagram illustrates the MABIC framework, which has two fusion approaches: Early Fusion and Late Fusion.}
    \label{fig:workflow}
\end{figure}

We introduce the \textbf{``MABIC''} framework, which combines unimodal text-based and multimodal-based classification approaches. The MABIC framework, which comprises the Early Fusion and Late Fusion fusion procedures, is depicted in Figure \ref{fig:workflow}. The codes and additional coding files for this research are available to the public at this link: \href{https://github.com/fatemafaria142/Uddessho-An-Benchmark-Dataset-for-Multimodal-Author-Intent-Classification-in-Bangla-Language}{Github}

\subsection{\textbf{Approach 1: Bangla Author Intent Text Classification}}

In this approach, we begin with text preprocessing to clean and standardize Bangla text, which involves tasks such as removing emojis, normalizing whitespaces, and stripping punctuation and special symbols for consistency. Next, we develop text-based intent classification models using state-of-the-art (SOTA) pre-trained language models like mBERT \cite{mbert}, DistilBERT \cite{Distil}, and XLM-RoBERTa \cite{XLM}. These models are fine-tuned on our dataset to leverage their advanced semantic comprehension capabilities. Following model development, we perform hyperparameter tuning with a batch size of 16, learning rate set to 0.001, and epochs ranging from 15 to 25 to optimize model performance. Finally, we evaluate our classification models, analyzing the results as presented in Table \ref{tab:performance_metrics} to gain valuable insights.

\subsection{\textbf{Approach 2: Multimodal Bangla Author Intent Classification}}
In order to comprehend author intent in Bangla content, a thorough examination of both textual and visual data is necessary. Our approach makes use of cutting-edge CNN architectures and pre-trained language models to accurately capture the complex patterns and representations required for intent classification. Our goal is to improve the classification model's accuracy by combining features from both modalities. We have divided our methodology into distinct steps to systematically address the various components of the classification process. 

{\textbf{Step 1) Text Features:}}
We utilize pre-trained language models such as mBERT, DistilBERT, and XLM-RoBERTa. These models are adept at capturing intricate linguistic patterns in Bangla content, providing a comprehensive representation of textual data.

{\textbf{Step 2) Image Features:}}
For visual representations, we employ eight different types of CNN architectures: ResNet 50 \cite{Resnet}, ResNet 101 \cite{Resnet}, ResNet 152 \cite{Resnet}, DenseNet 121 \cite{Dense}, DenseNet 169 \cite{Dense}, DenseNet 201 \cite{Dense}, MobileNet V2 \cite{moble}, and MobileNet V3 \cite{moble}. These CNNs enable us to extract detailed features from images, providing the dataset with valuable visual context.

{\textbf{Step 3) Fusion Techniques:}}
We integrate textual and visual data using Early Fusion \cite{EarlyFusion} and Late Fusion \cite{LateFusion} methods to enhance classification accuracy:

\begin{itemize}
    \item[$\bullet$] \textbf{Early Fusion}: This method combines text and image features at the input level, enabling the model to jointly learn from both modalities from the outset. This integrated learning process helps capture complex interactions between text and visual data. Mathematically, if $\mathbf{T}$ represents the text feature vector and $\mathbf{I}$ represents the image feature vector, Early Fusion can be expressed as:
 \begin{equation}
\mathbf{F}_{\text{early}} = f(\mathbf{W}_T \mathbf{T} + \mathbf{W}_I \mathbf{I} + \mathbf{b})
\end{equation}

    where $\mathbf{W}_T$ and $\mathbf{W}_I$ are weight matrices for text and image features respectively, $\mathbf{b}$ is the bias term, and $f$ is the activation function.\\

    \item[$\bullet$] \textbf{Late Fusion}: In this approach, text and image features are processed separately before being merged. This allows each modality to be independently refined, preserving their distinct nuances and contributing to a more nuanced final representation. If $\mathbf{T}$ and $\mathbf{I}$ are the independently processed text and image feature vectors, Late Fusion can be represented as:
\begin{equation}
\mathbf{F}_{\text{late}} = g(\mathbf{W}_T' h(\mathbf{T}) + \mathbf{W}_I' k(\mathbf{I}) + \mathbf{b}')
\end{equation}

    where $h(\mathbf{T})$ and $k(\mathbf{I})$ are the functions processing the text and image features respectively, $\mathbf{W}_T'$ and $\mathbf{W}_I'$ are weight matrices for the processed features, $\mathbf{b}'$ is the bias term, and $g$ is the activation function.
\end{itemize}

{\textbf{Step 4) Hyperparameter Tuning and Model Evaluation:}}
We adapt hyperparameters to optimize model performance and get the best results. Our approach uses the AdamW optimizer, a batch size of 16, and epochs of 30 to 60. The learning rate is adjusted between 0.001 and 0.01. Table \ref{tab:performance_metrics} lists measures we use to evaluate our Multimodal Bangla Author Intent Classification technique after hyperparameter tuning. The analysis helps modify and optimize the model, boosting accuracy.

\section{Result Analysis}
\subsection{Quantitative Analysis}

\begin{table}[htbp]
    \centering
    \caption{Detailed Performance Metrics of Various Pre-trained Language Models}
    \begin{tabular}{@{}c@{\hspace{10pt}}c@{\hspace{10pt}}c@{\hspace{10pt}}c@{\hspace{10pt}}c@{}}
    \toprule
    \textbf{Models} & \textbf{Accuracy} & \textbf{Precision} & \textbf{Recall} & \textbf{F1-Score} \\
    \midrule
    mBERT & 0.6300 & 0.6251 & 0.6324 & 0.6272 \\[2pt]
    \textbf{XLM-RoBERTa} & \textbf{0.6453} & \textbf{0.6418} & \textbf{0.6411} & \textbf{0.6415} \\[2pt]
    DistilBERT &  0.6016 & 0.5983 & 0.5898 &  0.5930 \\
    \bottomrule
    \end{tabular}
    \label{tab:performance_metrics}
\end{table}
In the unimodal approach Table \ref{tab:performance_metrics}, XLM-RoBERTa has the highest accuracy at 64.53\%, while DistilBERT has the lowest at 60.16\%.Table \ref{tab:multi} compares early and late fusion strategies in multimodal approaches. For the early fusion, the combination of ResNet50 and XLM-RoBERTa achieves the highest accuracy at 76.19\%, whereas ResNet101 and mBERT have the lowest accuracy at 63.58\%. In the late fusion approach, MobileNet V2 paired with mBERT reaches the highest accuracy of 75.55\%, with ResNet101 and XLM-RoBERTa gave the lowest at 62.67\%.

\begin{table}[htbp]
    \centering
    \caption{Comparative Performance Metrics of Early and Late Fusion Approaches Using Various Deep Learning Model Combinations}
    \begin{tabular}{ll>{\hspace{10pt}}c>{\hspace{10pt}}c>{\hspace{10pt}}c>{\hspace{10pt}}c>{\hspace{10pt}}c}
    \toprule
    \textbf{Approach} & \textbf{Models} & \textbf{Accuracy} & \textbf{Precision} & \textbf{Recall} & \textbf{F1-Score} \\ 
    \midrule                        
    \multirow{24}{*}{Early Fusion} & ResNet50 + mBERT & 0.7540 & 0.7500 & 0.7497 & 0.7488 \\
                                   & ResNet101 + mBERT & 0.6358 &	0.6420 & 0.6597 & 0.6376 \\
                                   & ResNet152 + mBERT & 0.7002 & 0.7040 & 0.7250 & 0.7120 \\
                                   & DenseNet121 + mBERT & 0.7560 & 0.7640 & 0.7580 & 0.7603 \\
                                   & DenseNet169 + mBERT & 0.7380 & 0.7455 & 0.7440 & 0.7333 \\
                                   & DenseNet201 + mBERT & 0.7476 & 0.7523 & 0.8220 & 0.7264 \\
                                   & MobileNet V2 + mBERT & 0.7170 &	0.7185 & 0.7770 &	0.7466 \\
                                   & MobileNet V3 + mBERT & 0.7476 & 0.7439 & 0.7434 & 0.7428 \\
                                   & \textbf{ResNet50 + XLM-RoBERTa} & \textbf{0.7619} & \textbf{0.7537} & \textbf{0.7636} & \textbf{0.7573} \\
                                   & ResNet101 + XLM-RoBERTa & 0.7340 & 0.71955 & 0.7359 & 0.7355 \\
                                   & ResNet152 + XLM-RoBERTa & 0.6624 & 0.6556 & 0.6562 & 0.6533 \\
                                   & DenseNet121 + XLM-RoBERTa & 0.6773 & 0.6730 & 0.6677 & 0.6694 \\
                                   & DenseNet169 + XLM-RoBERTa & 0.6506 & 0.6490 & 0.6480 & 0.6467 \\
                                   & DenseNet201 + XLM-RoBERTa & 0.6770 & 0.6185 & 0.6770 & 0.6741 \\
                                   & MobileNet V2 + XLM-RoBERTa & 0.7604 & 0.7642 & 0.7633 & 0.7579 \\
                                   & MobileNet V3 + XLM-RoBERTa & 0.6901 & 0.6843 & 0.7236 & 0.6948 \\
                                   & ResNet50 + DistilBERT & 0.7016 & 0.6986 & 0.7025 & 0.6933 \\
                                   & ResNet101 + DistilBERT & 0.6825 &	0.6840 & 0.6835 & 0.6765 \\
                                   & ResNet152 + DistilBERT & 0.6783 & 0.6717 & 0.6710 & 0.6679 \\
                                   & DenseNet121 + DistilBERT & 0.7170 & 0.7185 & 0.7770 & 0.7466 \\
                                   & DenseNet169 + DistilBERT & 0.6847 & 0.6892 & 0.6871 & 0.6829 \\
                                   & DenseNet201 + DistilBERT & 0.6454 & 0.6237 & 0.6300 & 0.6245 \\
                                   & MobileNet V2 + DistilBERT & 0.6550 & 0.6360 & 0.6366 & 0.6328 \\
                                   & MobileNet V3 + DistilBERT & 0.6901 & 0.6817 & 0.6870 & 0.6715 \\
    \midrule                
    \multirow{24}{*}{Late Fusion}  & ResNet50 + mBERT & 0.7188 & 0.7189 & 0.7173 & 0.7175 \\
                                   & ResNet101 + mBERT & 0.6773 & 0.6670 & 0.6630 & 0.6613 \\
                                   & ResNet152 + mBERT & 0.6581 & 0.6482 & 0.6484 & 0.6447 \\
                                   & DenseNet121 + mBERT & 0.6677 & 0.6637 & 0.6600 & 0.6620 \\
                                   & DenseNet169 + mBERT & 0.6698 & 0.6723 & 0.6746 & 0.6671 \\
                                   & DenseNet201 + mBERT & 0.6825 & 0.6890 & 0.6889 & 0.6795 \\
                                   & \textbf{MobileNet V2 + mBERT} & \textbf{0.7555} &	\textbf{0.7520} &	\textbf{0.7623} &	\textbf{0.7551} \\
                                   & MobileNet V3 + mBERT &0.6805 & 0.6840 & 0.6873 & 0.6838 \\
                                   & ResNet50 + XLM-RoBERTa & 0.7157 & 0.7167 & 0.7200 & 0.7166 \\
                                   & ResNet101 + XLM-RoBERTa & 0.6267 & 0.6445 & 0.6400 & 0.6344 \\
                                   & ResNet152 + XLM-RoBERTa & 0.7029 & 0.697 & 0.7124 & 0.7008 \\
                                   & DenseNet121 + XLM-RoBERTa & 0.7316 & 0.7281 & 0.7361 & 0.7300 \\
                                   & DenseNet169 + XLM-RoBERTa & 0.7412 & 0.7572 & 0.7455 & 0.7470 \\
                                   & DenseNet201 + XLM-RoBERTa & 0.6869 &	0.6846 &	0.6837	& 0.6836 \\
                                   & MobileNet V2 + XLM-RoBERTa & 0.7158 &	0.7267 & 0.7043 &	0.7153\\
                                   & MobileNet V3 + XLM-RoBERTa & 0.6645	& 0.6660 & 	0.7695& 	0.6634 \\
                                   & ResNet50 + DistilBERT & 0.6900 &	0.6944 &	0.6867 &	0.6883 \\
                                   & ResNet101 + DistilBERT & 0.6571 &	0.658 &	0.6536 & 0.6543 \\
                                   & ResNet152 + DistilBERT & 0.7492 &	0.7609 &	0.7539 &	0.7561 \\
                                   & DenseNet121 + DistilBERT & 0.7413 & 0.7390 & 0.7439 & 0.7407  \\
                                   & DenseNet169 + DistilBERT & 0.7438 &	0.7426 &	0.7520 &	0.7442\\
                                   & DenseNet201 + DistilBERT & 0.7380 &	0.7358 &	0.7464 &	0.7378\\
                                   & MobileNet V2 + DistilBERT & 0.7412 &	0.7394 &	0.7456 &	0.7401\\
                                   & MobileNet V3 + DistilBERT & 0.7492 &	0.7522 &	0.7523 &	0.7506\\
    \bottomrule
    \end{tabular}
    \label{tab:multi}
\end{table}
\subsection{Gaining Insight into Model Performance Through Error Analysis}
\begin{figure*}[htbp]
    \includegraphics[width=1\textwidth ]{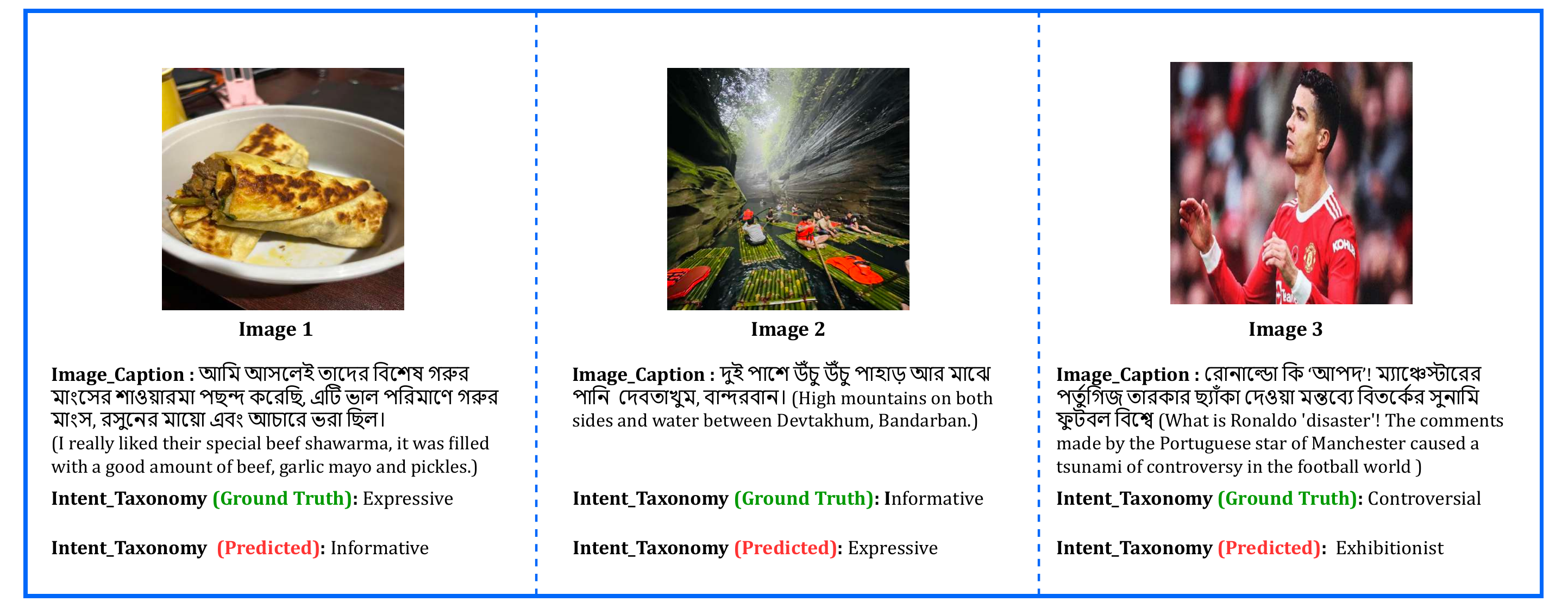}
    \caption{Error Analysis of MABIC framework results, illustrating both early fusion and late fusion techniques.}
    \label{fig:error}
\end{figure*}

In Figure \ref{fig:error}, the model predicted ``Informative'' for Image 1, focusing on factual details about shawarma. However, the post's emotional tone suggests misclassification, likely due to its descriptive emphasis over emotional content. Moving to Image 2, labeled ``Expressive,'' the model interpreted a description of Debotakhum in Bandarban as conveying enthusiasm. However, the primary intent was informative, highlighting the location's attributes rather than expressing personal feelings. Lastly, in Image 3, classified as ``Exhibitionist,'' the model focused on Ronaldo's provocative comments generating debate. Yet, the misclassification occurred because the model emphasized attention-seeking rather than recognizing the controversial nature of Ronaldo's comments.

\section{Limitations and Future Works}
Our research into understanding multimodal Bangla author intent has been limited by the complexities of Bangla language, such as dialects and variations, ambiguity in visuals, which impedes goal decision-making, and continuous uncertainty despite contextual cues from captions. In our future goal, we will investigate advanced multimodal fusion methods for integrating textual and visual data in order to improve intent identification. This includes trying out different transformer architectures to obtain insights. Furthermore, our primary focus will be on creating fine-grained taxonomies adapted to certain domains or circumstances, with the goal of improving accuracy. We will employ semiotic categories such as ``divergent,'' ``additive,'' and ``parallel'' to acquire more insights about user intent expression. Additionally, we will enhance model transparency and reliability by employing explainable AI techniques like GradCAM, GradCAM++, and ScoreCAM. 

\section{Conclusion}
In summary, our study represents a significant advancement in multimodal author intent classification, particularly within the Bangla language domain. Through the introduction of the ``Uddessho'' dataset and an examination of both unimodal and multimodal classification techniques, we achieved a 64.53\% accuracy in Bangla textual intent classification. Our proposed multimodal approach significantly outperformed this by achieving a 76.19\% accuracy. By employing cutting-edge language models and exploring fusion techniques for integrating textual and visual information, we have advanced author intent classification. Our findings not only offer valuable insights into understanding author intent through diverse modalities but also contribute to improving the accuracy of classification models. Moreover, our extensive evaluation of both text-based and multimodal classification approaches highlights the efficacy of incorporating visual features alongside textual data. While unimodal approaches provide substantial insights into author intent, our results demonstrate that the fusion of text and image features in multimodal approaches significantly enhances classification accuracy, offering an improvement of 11.66\% over traditional methods.

\end{document}